\documentclass{article}

\usepackage[preprint]{neurips_2025}

\usepackage[utf8]{inputenc} % allow utf-8 input
\usepackage[T1]{fontenc}    % use 8-bit T1 fonts
\usepackage{hyperref}       % hyperlinks
\usepackage{url}            % simple URL typesetting
\usepackage{booktabs}       % professional-quality tables
\usepackage{amsfonts}       % blackboard math symbols
\usepackage{nicefrac}       % compact symbols for 1/2, etc.
\usepackage{microtype}      % microtypography
\usepackage{xcolor}         % colors
\usepackage{mdframed}
\usepackage{tcolorbox}

\usepackage{amsmath}
\usepackage{globalvals}
\usepackage{graphicx}
\usepackage{wrapfig}
\usepackage{subfigure}
\usepackage{diagbox}
\usepackage{wrapfig}
\usepackage{subfigure}
\usepackage{multirow}
\usepackage{tabularx,colortbl}
\usepackage{soul}
\usepackage{float}

\defVal{model}{$\text{Re}^2$}

\newcommand{\vpara}[1]{\noindent\textbf{#1 }}
\title{\useVal{model}: A Consistency-ensured Dataset for Full-stage Peer Review and Multi-turn Rebuttal Discussions}

\author{%
  Daoze Zhang\textsuperscript{*}    \\
  Zhejiang University               \\
  \texttt{zhangdz@zju.edu.cn}       \\
  \And
  Zhijian Bao\textsuperscript{*}  \\
  Zhejiang University               \\
  \texttt{baozhijian@zju.edu.cn}  \\
  \And
  Sihang Du    \\
  Zhejiang University               \\
  \texttt{dusihang@zju.edu.cn}  \\
  \And
  Zhiyi Zhao \\
  Zhejiang University               \\
  \texttt{zhaozhiyi@zju.edu.cn}        \\
  \And
  Kuangling Zhang \\
  Zhejiang University               \\
  \texttt{zhangkl@zju.edu.cn}        \\
  \And
  Dezheng Bao \\
  Zhejiang University               \\
  \texttt{baodezheng@zju.edu.cn}        \\
  \And
  Yang Yang\textsuperscript{$\dagger$}      \\
  Zhejiang University               \\
  \texttt{yangya@zju.edu.cn}  \\
}

\begin{document}

\maketitle

\renewcommand{\thefootnote}{}
\footnotetext{\textsuperscript{*} Equal contribution. }
% \footnotetext{\textsuperscript{\Letter} Corresponding author.}
\footnotetext{\textsuperscript{$\dagger$} Corresponding author.}
\renewcommand{\thefootnote}{\arabic{footnote}}

\begin{abstract}

Peer review is a critical component of scientific progress in the fields like AI, but the rapid increase in submission volume has strained the reviewing system, which inevitably leads to reviewer shortages and declines review quality. 
Besides the growing research popularity, another key factor in this overload is the repeated resubmission of substandard manuscripts, largely due to the lack of effective tools for authors to self-evaluate their work before submission. 
Large Language Models (LLMs) show great promise in assisting both authors and reviewers, 
and their performance is fundamentally limited by the quality of the peer review data.
% but their performance ceiling heavily depends on high-quality peer review data.
However, existing peer review datasets face three major limitations: 
(1) limited data diversity, (2) inconsistent and low-quality data due to the use of revised rather than initial submissions, and (3) insufficient support for tasks involving rebuttal and reviewer-author interactions. 
To address these challenges, we introduce the largest consistency-ensured peer review and rebuttal dataset named \useVal{model}, which comprises 19,926 initial submissions, 70,668 review comments, and 53,818 rebuttals from 24 conferences and 21 workshops on OpenReview.
Moreover, the rebuttal and discussion stage is framed as a multi-turn conversation paradigm to support both traditional static review tasks and dynamic interactive LLM assistants, providing more practical guidance for authors to refine their manuscripts and helping alleviate the growing review burden.
Our data and code are available in 
\href{https://anonymous.4open.science/r/ReviewBench_anon/}{this repository}.

% This provides more practical guidance for authors to refine their manuscripts and helps alleviate the growing review burden in AI research.

% offering more practical guidance for authors to refine submissions and ease the review burden in AI research.

% helping authors refine submissions and easing the review burden in AI research.

% which enables the training and evaluation of dynamic, interactive LLM-based reviewing assistants, 
% it includes structured multi-turn conversations, ensures consistency by using only initial submissions, and unifies heterogeneous data formats. 
% This dataset is the largest and most comprehensive of its kind, enabling the development of LLMs with enhanced capabilities for review generation, rebuttal handling, and interactive author support.

% To our knowledge, \useVal{model} is the largest real-world peer review dataset, with the widest conference coverage and most complete review stage data. It is also the first to ensure consistency for multi-turn rebuttal tasks. \useVal{model}

\end{abstract}
\section{Introduction}

Peer review is a cornerstone in the advancement of scientific research, ensuring that high-value works which are sufficiently novel, credible, and rigorously evaluated can be published. 
With the rapid surge in submission volume in some fields like Computer Science (CS) or Artificial Intelligence (AI), 
peer researchers have to bear increasing pressure to review, leading to a shortage of reviewers and an inevitable decline in review quality.
In addition to the growing popularity of these disciplines, another major contributor to the submission overload is the repeated resubmission of manuscripts that fall short of quality standards. 
This trend is often driven by the lack of effective tools for authors to objectively assess and improve their manuscripts prior to submission, resulting in multiple rounds of submission for the same study.
For the above two issues in the research and review stages,
Large Language Models (LLMs) hold significant potential to become a powerful assistant to alleviate these problems.
Specifically, LLM-based review tools can be used by authors as a pre-submission self-evaluation mechanism, allowing authors to identify and improve the weaknesses in the manuscripts, and thereby enhancing the work quality and reducing resubmission rates. 
Additionally, LLMs can be directly integrated into the peer review process to support reviewers in generating more specific and constructive feedback — as demonstrated by the ICLR 2025 conference, which has introduced an LLM-powered reviewer assisting system~\citep{iclr2025assist}.

For the training of language models with peer review capabilities, 
the most crucial knowledge foundation that determines the upper bound of the model performance is the real-world high-quality paper review data.
Numerous researchers have constructed diverse peer review datasets to support the training and evaluation of LLM-based reviewing assistants. 
These existing datasets target a range of tasks, including: 
(1) prediction-oriented tasks such as acceptance prediction and score prediction ~\citep{kang2018dataset,bharti2021peerassist,dycke2022nlpeer}; 
(2) generation tasks, including review and meta-review generation~\citep{shen2021mred,zhang2022investigating,yuan2022can,wu2022incorporating,d2023aries,jin2024agentreview,gao2024reviewer2,zhou2024llm,weng2024cycleresearcher,zhu2025deepreview}; and 
(3) analytical tasks related to review content, like review action analysis~\citep{kennard2021disapere,purkayastha2023exploring,bharti2024politepeer}.
Despite the above advancements in the field of automated paper review, existing datasets still exhibit the following critical limitations.

From the perspective of data diversity, \textbf{the paper sources of these existing datasets are limited to a few conferences, and the amount of data is also relatively small.}
Due to the huge differences in the availability and formatting of the review information across different conferences, 
the majority of existing datasets~\citep{kennard2021disapere,bharti2021peerassist,wu2022incorporating,zhang2022investigating,zhou2024llm,weng2024cycleresearcher,jin2024agentreview,zhu2025deepreview} are only based on the review data from the ICLR conference, which is known for its high level of transparency in the review process. 
% thereby avoiding the problem of large differences in data formats between different conferences. 
A few other works~\citep{kang2018dataset,shen2021mred,yuan2022can,dycke2022nlpeer,d2023aries,gao2024reviewer2} incorporate reviews from additional conferences like NeurIPS and ACL. 
However, none of them comprehensively capture all publicly available peer reviews from conferences hosted on OpenReview, leading to obvious limitations in both the data source diversity and data scale.

From the viewpoint of data quality, most existing datasets have a fatal problem — \textbf{the provided paper content may correspond to the revised version rather than the initial submission.} 
Obviously, for training and evaluating language models on the task of automated paper review, the paper content fed into models is supposed to be the initial submission that is not revised by the author, as this is the version to which the reviewers’ comments are actually addressed.
However, as shown in Tab.~\ref{tab:works_compare}, most existing datasets (not highlighted in yellow) fail to guarantee that the paper contents are indeed the initial submissions,
% and are highly likely to contain author revisions. 
and contain versions that have been revised by the authors in response to review comments.
This discrepancy introduces substantial risks to the coherence and consistency of the review data, undermining the rationality of model training and evaluation for the review-related tasks.

From the task perspective, many existing works mainly \textbf{stay on the traditional review generation task, while overlooking the valuable data contained in the rebuttal and discussion stages.}
In order to better assist authors in self-evaluation and improving their submissions before formal review, language models are expected not only to generate static review comments, but also to understand the author's rebuttals and provide further responses. 
This rebuttal–discussion process constitutes a typical multi-turn conversation task, which holds great potential and research value. 
However, as shown in Tab.~\ref{tab:works_compare}, most existing datasets do not contain any information related to rebuttals. 
For the few datasets~\citep{kennard2021disapere,wu2022incorporating,jin2024agentreview,tan2024peer} involving rebuttals, 
they often handle the rebuttal and discussion data in a coarse and insufficient manner and also fail to guarantee the consistency of the data,
% their handling of rebuttal and discussion data is often relatively coarse and insufficient, 
falling far short of supporting research that treats rebuttal and discussion as a well-defined multi-turn conversation task.

To tackle the limitations above, we propose a real-world dataset named \useVal{model} for comprehensive academic peer review, to support the training and evaluation of both \textbf{re}view and \textbf{re}buttal abilities of language models.
Our \useVal{model} dataset consists of two main parts:
(1) the \useVal{model}-Review dataset contains 19,926 initial paper submissions and 70,668 review comments from human reviewers collected from OpenReview, covering 24 conferences and 21 workshops from 2017 to 2025; and 
(2) the \useVal{model}-Rebuttal dataset contains 14,830 initial submissions paired with 53,818 rebuttal and discussions, which are formatted as structured multi-turn conversation to facilitate the training of various language models.
By overcoming the huge heterogeneity in data storage formats across different conferences and years, we unify all data into a consistent format to facilitate broad accessibility for peer researchers.
Moreover, our \useVal{model} dataset covers as many review stages on OpenReview as possible, including  initial submissions, reviewer comments, ratings and confidence scores, aspect-specific ratings (e.g., soundness, presentation, contribution), rebuttal–discussion conversations, score changes before and after rebuttal, meta-reviews, and final decisions.

To our knowledge, our \useVal{model} is the largest real-world peer review dataset to date, with the broadest inclusion of conferences and the most comprehensive coverage of review stages (details in Tab.~\ref{tab:works_compare}).
In addition, it is the first consistency-ensured dataset to support rebuttal tasks in a multi-turn conversation paradigm. 
The \useVal{model} dataset not only allows traditional static tasks such as computational analysis, acceptance or score prediction, review or meta-review generation, but also enables the training of interactive, chat-based models for further rebuttal and discussion, offering support for authors to self-evaluate and improve their work before submission, which is also helpful in reducing the review burden in the AI community.
In summary, our key contributions are as follows:

\begin{itemize}
    \item 
    To our knowledge, we present the largest real-world consistency-ensured dataset named \useVal{model} for peer review and rebuttal-discussion, which features the widest range of conferences and the most complete review stages, including initial submissions, reviews, (aspect) ratings and confidence, rebuttals, discussions, score changes, meta-reviews, and decisions.
    \item 
    Moving beyond the traditional static review paradigm, we treat the rebuttal and discussion data as a multi-turn conversation task between reviewers and authors, which enables the training and evaluation of dynamic, interactive LLM-based reviewing assistants, offering more practical guidance for authors to self-improve their work before submission.
    \item 
    We conduct a statistical analysis of the proposed dataset, and experimentally demonstrate its effectiveness in improving the capabilities of language models in peer review and rebuttal scenarios through four review-related tasks.
    % Experimental results demonstrate that ... models trained on \useVal{model} significantly outperform prior baselines on review generation and rebuttal understanding tasks...
\end{itemize}

\begin{table}[h]
\centering
\caption{ \label{tab:works_compare}
    \textbf{Comparison between our \useVal{model} dataset and existing peer review datasets.}
    Note that only the datasets in yellow rows can guarantee all provided papers are \colorbox{yellow}{initial submissions}, which is critical for the consistency and quality of the review data.
    For the ``Task'' column, AP is the abbreviation for Acceptance Prediction, SP for Score Prediction, RA refers to Review Analysis tasks,
    % (e.g., review action analysis or fairness studies), 
    RG stands for Review Generation, and MG denotes Meta-review Generation.
    In the three columns representing numbers, ``-'' means please refer to the original dataset for the number. 
    % The blank spaces denote that the dataset does not contain this item.
}

\vspace{-1mm}

% \scriptsize
\footnotesize
\setlength{\tabcolsep}{1.7mm} %列距离
{
\renewcommand{\arraystretch}{0.88} %行距离
\setlength{\arrayrulewidth}{0.3pt}  % \hline的线粗
\begin{tabular}{lcccccccccccccc}
\toprule
% & & & \multicolumn{7}{c}{Intermediate Review Stages}  & \multicolumn{4}{c}{\#Sample}  & \multicolumn{1}{c}{\multirow{2}{*}{Real-world or Synthetic}} \\
\rotatebox{90}{Dataset Name} & \rotatebox{90}{\begin{tabular}[c]{@{}l@{}}Data Source \\ (Conf.\&Year) \end{tabular}}  & \rotatebox{90}{Task} & \rotatebox{90}{Reviews} & \rotatebox{90}{Aspect Scores} & \rotatebox{90}{Rebuttal} & \rotatebox{90}{Score Changes} & \rotatebox{90}{Meta-reviews} & \rotatebox{90}{Final Decision} & \rotatebox{90}{\# Paper} & \rotatebox{90}{\begin{tabular}[c]{@{}l@{}} \# Review \\ Comments \end{tabular}} & \rotatebox{90}{\# Rebuttal} \\
\midrule
\begin{tabular}[c]{@{}l@{}}PeerRead \\ \scriptsize\citep{kang2018dataset} \end{tabular} & 
 \begin{tabular}[c]{@{}c@{}} ICLR 17, ACL 17, \\ NeurIPS 13-17 \end{tabular} & AP, SP & \checkmark & \checkmark &   &    & \checkmark & \checkmark & 14,700   & 10,700 & 0 \\
% \hline
\cmidrule(lr){1-12}
\begin{tabular}[c]{@{}l@{}}DISAPERE \\ \scriptsize\citep{kennard2021disapere} \end{tabular}   & ICLR 19-20 & RA   & \checkmark &   & \checkmark &    &   &   & 0 & 506 &  506   \\
% \hline
\cmidrule(lr){1-12}
\begin{tabular}[c]{@{}l@{}} PEERAssist \\ \scriptsize\citep{bharti2021peerassist} \end{tabular} & ICLR 17-20   & AP & \checkmark &   &   &    &   & \checkmark & 4,467& 13,401 & 0 \\
% \hline
\cmidrule(lr){1-12}
\rowcolor{yellow}
\begin{tabular}[c]{@{}l@{}} MReD \\ \scriptsize\citep{shen2021mred} \end{tabular} & ICLR 18-21 & MG & \checkmark &   &   &    &   & \checkmark & 7,894 & 30,764 & 0  \\
% \hline
\cmidrule(lr){1-12}
\begin{tabular}[c]{@{}l@{}}ASAP-Review \\ \scriptsize\citep{yuan2022can} \end{tabular} & \begin{tabular}[c]{@{}c@{}} ICLR 17-20, \\ NeurIPS 16-19 \end{tabular}  & RG & \checkmark &   &   &    & \checkmark &   & 8,877& 28,119 & 0    \\
% \hline
\cmidrule(lr){1-12}
\rowcolor{yellow}
\begin{tabular}[c]{@{}l@{}} NLPeer \\ \scriptsize\citep{dycke2022nlpeer} \end{tabular} & \scriptsize\begin{tabular}[c]{@{}c@{}} ACL 17, ARR 22, \\ COLING 20, \\ CONLL 16 \end{tabular} & RA & \checkmark & \checkmark &   &    &   &   & 5,672 & 11,515 & 0   \\
% \hline
\cmidrule(lr){1-12}
\begin{tabular}[c]{@{}l@{}} PRRCA \\ \scriptsize\citep{wu2022incorporating} \end{tabular}  & ICLR 17-21   & MG & \checkmark &   & \checkmark & \checkmark  & \checkmark & \checkmark & 7,627 & 25,316 & -  \\
% \hline
\cmidrule(lr){1-12}
\scriptsize\citep{zhang2022investigating} & ICLR 17-22   & RA & \checkmark & \checkmark &   &    &   & \checkmark & 10,289   & 36,453 & 68,721 \\
% \hline
\cmidrule(lr){1-12}
\rowcolor{yellow}
\begin{tabular}[c]{@{}l@{}} ARIES \\ \scriptsize\citep{d2023aries} \end{tabular}  & OpenReview & RG & \checkmark &   &   &    &   &   & 1,720 & 4,088 & 0 \\
% \hline
\cmidrule(lr){1-12}
\begin{tabular}[c]{@{}l@{}}AgentReview \\ \scriptsize\citep{jin2024agentreview} \end{tabular} & ICLR 20-23   & RG, MG & \checkmark & \checkmark & \checkmark & \checkmark  & \checkmark & \checkmark & 500  & 10,460 &  - \\
% \hline
\cmidrule(lr){1-12}
\begin{tabular}[c]{@{}l@{}} Reviewer2 \\ \scriptsize\citep{gao2024reviewer2} \end{tabular} & \scriptsize \begin{tabular}[c]{@{}c@{}} ICLR 17-23, \\ NeurIPS 16-22 \\ (PeerRead \& NLPeer) \end{tabular} & RG & \checkmark &   &   &    & \checkmark & \checkmark & 27,805   & 99,727 & 0   \\
% \hline
\cmidrule(lr){1-12}
\begin{tabular}[c]{@{}l@{}} RR-MCQ \\ \scriptsize\citep{zhou2024llm} \end{tabular} & \begin{tabular}[c]{@{}c@{}}ICLR 17 \\ \scriptsize(from PeerRead) \end{tabular}   & RG & \checkmark &   &   &    &   & \checkmark & 14& 55 & 0   \\
% \hline
\cmidrule(lr){1-12}
\begin{tabular}[c]{@{}l@{}}ReviewMT \\ \scriptsize\citep{tan2024peer} \end{tabular} & ICLR 17-24 & RG & \checkmark &  & \checkmark & & \checkmark & \checkmark & 26,841 & 92,017 & 0 \\
% \hline
\cmidrule(lr){1-12}
\begin{tabular}[c]{@{}l@{}}Review-5K \\ \scriptsize\citep{weng2024cycleresearcher} \end{tabular}  & ICLR 24  & RG & \checkmark &   &   &    &   & \checkmark & 4,991 & 16,000 & 0 \\
% \hline
\cmidrule(lr){1-12}
\begin{tabular}[c]{@{}l@{}}DeepReview-13K \\ \scriptsize\citep{zhu2025deepreview} \end{tabular}    & ICLR 24-25   & RG & \checkmark & \checkmark &   &    &   & \checkmark & 13,378   & 13,378 & 0  \\
\midrule
\rowcolor{yellow}
\textbf{Our \useVal{model}} & \begin{tabular}[c]{@{}c@{}} 45 venues \\ from 2017 to 2025 \end{tabular} & \begin{tabular}[c]{@{}c@{}} RG, MG, \\ AP, SP\end{tabular} & \checkmark & \checkmark & \checkmark & \checkmark  & \checkmark & \checkmark & \textbf{19,926} & \textbf{70,668} & \textbf{53,818} \\
\bottomrule
\end{tabular}
}

\vspace{-6mm}

\end{table}

\section{The \useVal{model} Dataset}

\subsection{Data Collection and Processing}

\vpara{\useVal{model}-Review Dataset.}
The first subset of our proposed dataset is named \useVal{model}-Review, which is mainly designed for acceptance prediction, score prediction, and review or meta-review generation tasks. 
For the \useVal{model}-Review dataset, 
% our data collection and processing pipeline is as follows.
we first utilize the official API of OpenReview to automatically crawl all publicly accessible papers and their full peer review records (including metadata, reviews, rebuttals, etc.) from OpenReview, covering 68 conferences from 2013 to 2025.
Afterwards, given that all the review-related tasks must be grounded in the initial submitted manuscripts rather than revised versions, we need to ensure that the paper contents in our dataset are the initial submissions.
% filter the crawled data to retain only those papers for which the original submissions were publicly available. 
To achieve this, we comprehensively collect the paper submission deadlines for each conference of different years. 
Based on the information of deadlines, we then employ web scraping techniques to extract the latest version before the submission deadline from the ``Revision History'' page of each paper.
The final paper set spans 24 conferences and 21 workshops from 2017 onward.

For the extraction of review contents, due to the diversity of conferences and years, these review data come in a wide range of formats. 
Therefore, the processing logic in existing works, which typically target common conferences such as ICLR or NeurIPS, cannot be directly applied to these review data from all the 45 venues. 
To address this challenge, we manually audit all data format variations involved, and implement customized extraction logic for each conference-year pair, to achieve automatic and efficient extraction of full-stage review contents across these conferences.

For the paper content, we convert paper formats from PDF into plain text to facilitate the use by the research community.
To achieve this, we employ a commercial tool named Doc2X\footnote{\url{https://github.com/NoEdgeAI/doc2x-doc}}, which outperforms open-source alternatives in terms of recognition accuracy and quality, especially for mathematical formulas. 
Using this tool, we convert the paper content in our dataset into both LaTeX and Markdown formats for broader accessibility and downstream applications.

\begin{figure}[ht]
\begin{center}

\includegraphics[width=\linewidth]{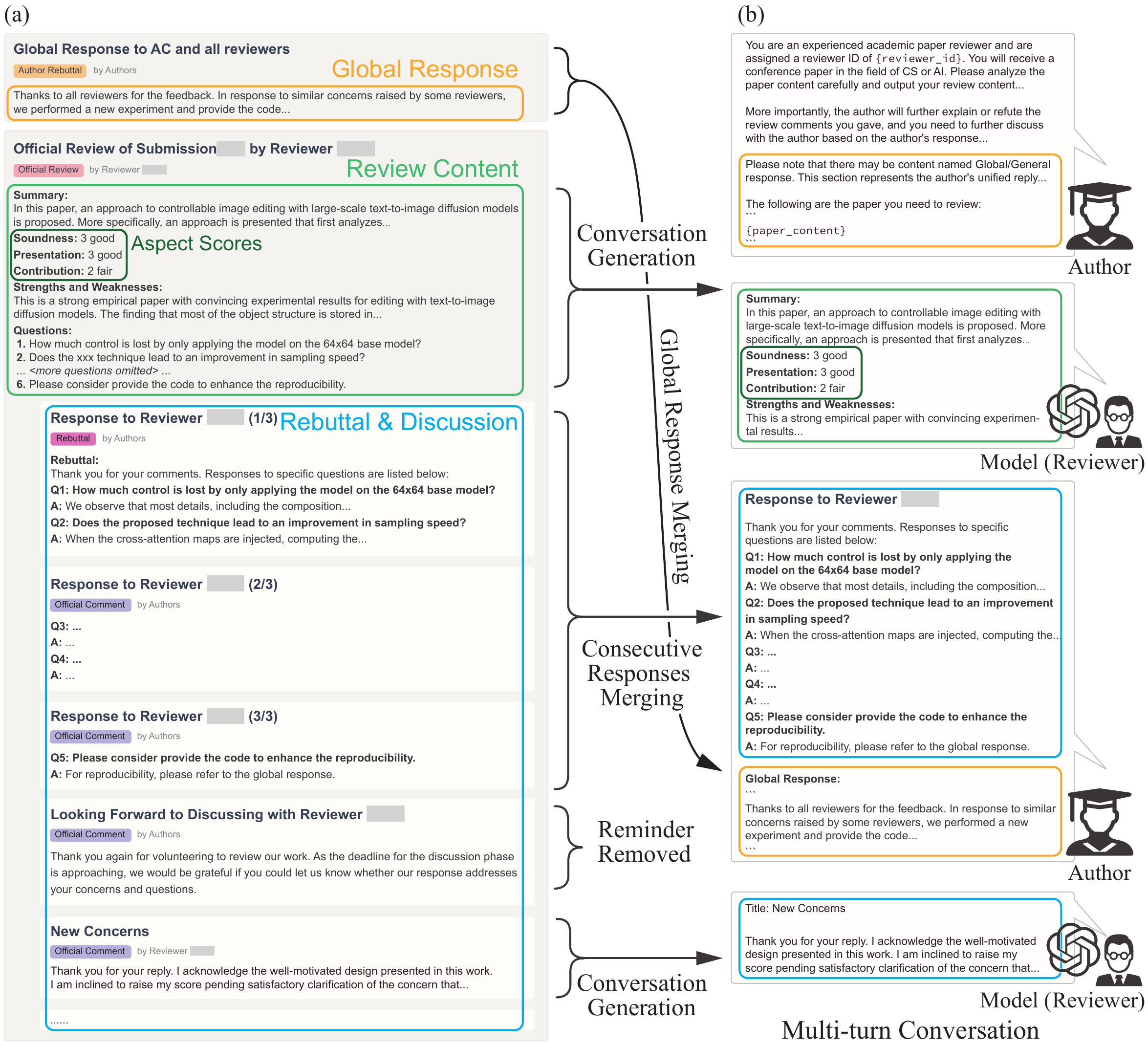}

\vspace{-2mm}

\caption{
    \label{fig:rebuttal_pipeline}
    \textbf{The conversion from raw review and rebuttal data to multi-turn conversations.}
    For the raw review data crawled from OpenReview (as shown in sub-figure(a)), we concatenate multiple consecutive responses from the same role (author or reviewer) into a single turn.
    In cases where the author's final response is merely a reminder or urging, we adopt a hybrid strategy combining manual inspection and automated methods to identify and remove such reminder responses.
    As for the global responses,
    % provided by authors to address common concerns raised by multiple reviewers, 
    we insert them into the dialogue at the appropriate position, treating it as supplementary reference rather than direct conversation content. 
    Finally, as shown in sub-figure(b), we construct a self-consistent, high-quality, and information-complete multi-turn conversation dataset.
}

\end{center}

\vspace{-3mm}

\end{figure}

\vpara{\useVal{model}-Rebuttal Dataset.}
Another subset named \useVal{model}-Rebuttal is designed as a structured multi-turn conversation dataset based on the rebuttal and discussion data between authors and reviewers.
For the construction of our \useVal{model}-Rebuttal subset, 
we further filter the review dataset mentioned above to retain only those papers for which the reviewer–author rebuttal stage is publicly accessible, 
and then process the corresponding rebuttal interactions. 
Specifically, we organize the authors' rebuttals, along with the subsequent reviewer–author discussion, into structured multi-turn dialogues.
The main challenges encountered here are mainly the following two parts.

First, due to the character limits imposed on each individual response on OpenReview, authors often post multiple consecutive responses during the rebuttal stage, with each response containing only a portion of their overall response (see the blue box in Fig.~\ref{fig:rebuttal_pipeline}(a) for an example).
To convert the rebuttal–discussion process into a well-structured multi-turn dialogue for the training of language models, we concatenate multiple consecutive responses from the same role (e.g., author or reviewer) into a single turn. 
Then, DeepSeek-R1~\citep{guo2025deepseek} is further employed to merge the title of each response, producing a coherent full response along with a unified title.
It is worth noting that, in some cases, the final response in a series of consecutive posts from the author is simply a reminder or urging to an unresponsive reviewer. 
Clearly, such responses should not be concatenated with the previous rebuttal content. 
To handle this problem, we adopt a hybrid strategy combining manual inspection and automated methods to identify and exclude these follow-up reminders, ensuring the generated multi-turn dialogue data is consistent and high-quality.

Second, when reviewing a submission, sometimes several reviewers may raise similar concerns, and the author will give a unified response to these shared questions through a global or general response (see the orange box in Fig.~\ref{fig:rebuttal_pipeline}(a) for an example).
To construct a complete multi-turn rebuttal–discussion conversation, we incorporate the global responses into the author–reviewer interactions. 
Specifically, we insert the content of global responses into the dialogue at the appropriate position corresponding to each reviewer’s related comment.
However, since global responses may also address concerns raised by other reviewers not involved in the current dialogue thread, we need to take special care to maintain logical consistency. 
Therefore, we insert global responses using a special format (as illustrated in the orange box in Fig.~\ref{fig:rebuttal_pipeline}(b)), treating them as referential context instead of direct replies from the author within the turn. 
This supplements the dialogue with reference information while preserving the consistency and logical flow of the multi-turn conversation.

\subsection{Dataset Statistics}

\begin{figure}[ht]
\begin{center}

\includegraphics[width=\linewidth]{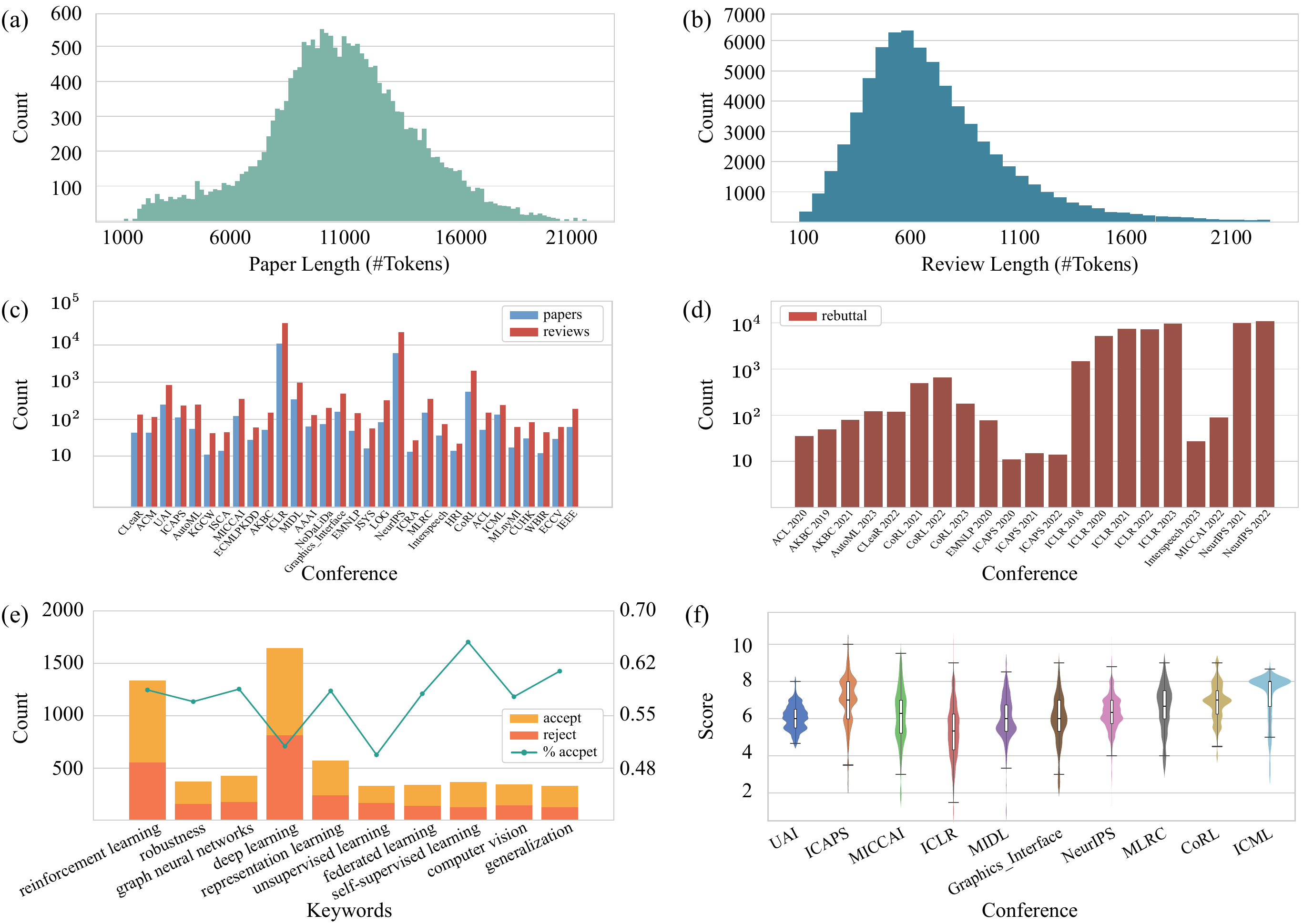}

\vspace{-1mm}

\caption{
    \textbf{Statistics of our \useVal{model} dataset.}
    (a) Distribution of the length of papers in tokens.
    (b) Distribution of the length of reviews in tokens.
    (c) Distribution of the number of papers and reviews in each conference.
    (d) Distribution of the number of rebuttals in each conference of each year.
    (e) Submission counts and acceptance proportion across the 10 most frequent keywords.
    (f) Violin plot (with a box plot inside) of review scores across the top 10 conferences with the most papers.
}
\label{fig:statistic_dataset}
\end{center}

\vspace{-1mm}

\end{figure}

% Our \useVal{model} dataset includes the latest papers before the review, metadata, metareview, decision, reviews, rebuttals, and more. 
%Here we provide the statistical details of our \useVal{model} dataset.
% and show the distribution of the key data.

% \vpara{Basic Statistics.}
\vpara{Paper, Review, and Rebuttal Distributions.}
% Our \useVal{model} dataset contains a total of 19,926 papers, 70,668 reviews, and 53,818 rebuttals. 
The distributions of paper and review lengths (in number of tokens) are shown in the histograms in Fig.~\ref{fig:statistic_dataset}(a) and Fig.~\ref{fig:statistic_dataset}(b), respectively.
Most papers range in length from 6,000 to 16,000 tokens, 
while the majority of reviews are distributed between 300 and 1,100 tokens. 
As shown in Fig.~\ref{fig:statistic_dataset}(c), we count the total number of papers and reviews of each conference. 
Among them, ICLR and NeurIPS hold the highest proportion of the number of papers and reviews.
% Since not every conference has rebuttals every year, we filter all conference in the \useVal{model} dataset. 
The distribution of the number of rebuttals for each conference in each year is given in Fig.~\ref{fig:statistic_dataset}(d), 
where ICLR and NeurIPS again account for the highest proportion.
% Notably, most reviews of ICLR and NeurIPS have multiple rounds of rebuttals.

\vpara{Acceptance and Scores.}
% In this part, we conduct an in-depth analysis of the \useVal{model} dataset. 
As shown in Fig.~\ref{fig:statistic_dataset}(e), to examine the popularity of different research areas within the AI community,
we present the number of papers and the acceptance proportions across the top 10 frequent research keywords. 
It can be illustrated that the acceptance proportions of ``self-supervised learning'' and ``generalization'' rank first and second.
Meanwhile, the number of papers about ``deep learning'' and ``reinforcement learning'' is the largest, demonstrating that these two are popular research areas in the field of AI. 
Also, in order to present the distributions of the review scores among different conferences, we utilize the violin plot with a box plot inside, to conduct a statistical analysis of the normalized scores of each conference in Fig.~\ref{fig:statistic_dataset}(f). 
Due to the excessive number of conferences, we only show the top 10 conferences with the most papers in detail. 
% With the aim of presenting it intuitively, we uniformly map the scores of each conference to 0-10. 
In the violins of Fig.~\ref{fig:statistic_dataset}(f), the bold horizontal line inside the box represents the median score, and the contour width indicates the number of papers of this score. 
It is worth noting that since Kernel Density Estimation (KDE) is used to smooth the data distribution, it may cause the tail extension of the violin to exceed the range boundary. 
It can be demonstrated that the median scores of each conference are all between 5 and 7. 
Additionally, by counting the peaks in each violin, we find that ICLR has almost no peaks, indicating that the score distribution of the ICLR submissions is the most uniform.
More details about the violin plot are explained in App.~\ref{app:violin_plot}.

\section{Experiment}

\subsection{Experimental Setup}

To demonstrate the versatility of our dataset, we conduct experiments on four tasks related to paper peer review, including acceptance prediction, score prediction, review generation, and rebuttal-discussion conversation.
For the \useVal{model}-Review dataset, we sample 1,000 papers along with their reviews as the test set, with the remaining data used for training. 
For the \useVal{model}-Rebuttal dataset, 500 papers and the rebuttals are selected for testing, and the rest are used for training.
We employ several open-source LLMs to evaluate their performance on our datasets, including LLaMA-3.1-8B-Instruct~\citep{meta2025introducing} and Qwen2.5-7B-Instruct~\citep{yang2024qwen2_5}. The LLaMA-3.1-8B-Instruct and Qwen2.5-7B-Instruct are fine-tuned on our training data using LoRA~\citep{hu2022lora} for one epoch on a learning rate of $1 \times 10^{-4}$ with cosine scheduler, and we report both the fine-tuned and zero-shot results. 
In addition, for the prediction and review generation tasks, we further conduct experiments with other models specifically designed for peer review scenarios, including SEA-E~\citep{yu2024automated}, LLaMA-OpenReviewer-8B~\citep{idahl2024openreviewer}, DeepReviewer-14B~\citep{zhu2025deepreview}, and CycleReviewer-8B~\citep{weng2024cycleresearcher}
(details in App.~\ref{app:baselines}).
All the training and evaluation are conducted on four NVIDIA A100 80G GPUs.
The details of the four tasks are as follows:

\vpara{Acceptance and Score Prediction. }
The acceptance prediction task aims to predict whether a paper will be accepted or rejected based on its content. It is a two-class classification task (acceptance or rejection), so we use accuracy, precision, recall, and F1 score as the evaluation metrics of the acceptance prediction task. 
Further, score prediction focuses not just on acceptance outcomes, but on predicting the detailed review scores (e.g., overall rating) that a paper would receive. 
Its performance is typically measured using Mean Absolute Error (MAE) and Mean Squared Error (MSE). 

\vpara{Review Generation. }
The review generation task is to automatically generate peer reviews for the given paper content, simulating the feedback provided by human reviewers.
Common evaluation metrics for this task include BLEU~\citep{papineni2002bleu} and ROUGE~\citep{lin2004rouge}, which assess the lexical overlap between generated reviews and reference reviews.
In addition to the traditional metrics above, we also employ two embedding-based metrics, including the BERTScore and the cosine similarity of embedding (EmbedCos). 
% Specifically, we utilize a 12-layer DeBERTa-large-MNLI model to calculate the F1 score as the final BERTScore metric. 
% For the EmbedCos, we use the all-mpnet-base-v2 model to encode the texts and compute their embedding similarities, providing a holistic measure of semantic similarity beyond surface-level matching.
Specifically, we adopt a 12-layer 
% \texttt{microsoft/deberta-large-mnli}
DeBERTa-large-MNLI
~\citep{he2020deberta} for BERTScore and the \texttt{sentence-transformers/all-mpnet-base-v2} model for EmbedCos.
% Furthermore, LLaMA-3.1-8B-Instruct is used as an evaluator to conduct qualitative assessments of review generation outputs.
More details of these evaluation metrics are given in App.~\ref{app:eval_metrics}.

\vpara{Rebuttal-Discussion Conversation. }
This task aims to simulate reviewer-author interactions during the rebuttal stage of peer review, providing coherent, context-aware, and constructive feedback across turns.
% In this setting, the model acts as a reviewer engaging in multiple rounds of discussion based on an author’s initial rebuttal. 
The basic metrics we use include BLEU, ROUGE, BERTScore, and embedding similarity.
% as well as politeness, persuasiveness, and specificity to capture the quality of rebuttal discourse. 
Also, to more deeply and thoroughly evaluate the ability of language models to simulate reviewer–author discussions, 
we further employ the LLaMA-3.1-8B-Instruct as a judge to evaluate model responses across five specific dimensions, including accuracy, constructiveness, completeness, clarity, and quality.
More explanations about these five aspects and the judge instruction are given in App.~\ref{app:rebuttal_llm_judge}.

\subsection{Experimental Results}

\vpara{Acceptance and Score Prediction. }
As shown in Tab.~\ref{tab:review_exp_pred}, 
SEA-E and DeepReviewer-14B achieve strong results in accuracy, recall, and F1 score, indicating accurate judgment of paper quality and acceptance boundaries. 
In contrast, LLaMA-OpenReviewer-8B and CycleReviewer-8B have higher precision but lower F1, suggesting they are more strict and conservative in accepting papers.
For the open-source LLMs, LLaMA-3.1-8B and Qwen2.5-7B accept all papers without discrimination, reflecting a common flaw in LLMs: a tendency to please humans without criticism. 
Therefore, in the table we color them grey to indicate that the result is useless, and exclude them from the bold and underline marking of the first and second place.
After finetuning, both models show more reasonable outputs, with notable decreases in MAE and MSE for score prediction, demonstrating that the finetuning on our training data significantly improves the review ability.
These changes highlight the effectiveness of our dataset in enhancing models' peer-review capabilities, enabling them to better capture the judgment basis and patterns of human reviewers.

\vpara{Review Generation. }
As shown in Tab.~\ref{tab:review_exp_gene}, the fine-tuned LLaMA-3.1-8B ranks first among all models in both BLEU and ROUGE-L metrics, greatly outperforming its zero-shot version, indicating that the fine-tuning on our dataset allows LLaMA-3.1-8B to better match the language structure and phrasing typical of real-world peer reviews.
In contrast, the results of DeepReviewer-14B and CycleReviewer-8B are obviously low on BLEU and ROUGE-L, because their generated content emphasizes abstract review reasoning rather than direct response.
In terms of the EmbedCos, which evaluates similarity between semantic vectors, fine-tuned LLaMA-3.1-8B and Qwen2.5-7B achieve substantial improvements of 49.9\% and 20.5\% compared with the third place, respectively, suggesting a high semantic alignment between their generated texts and real reviews in the embedding space.
Therefore, the training on our dataset significantly boosts models’ review generation capabilities, laying a solid foundation for building language model-based peer review assistants.

\newcommand{\bad}[1]{{\color{gray}  #1}}

\begin{table}[ht]
\centering
\caption{ \label{tab:review_exp_pred}
    Results comparison between open-source LLMs and baselines on prediction tasks.
}

\vspace{-2mm}

% \scriptsize
\footnotesize
\setlength{\tabcolsep}{1.7mm} %列距离
{

\begin{tabular}{lrrrrrr}
\toprule
\multicolumn{1}{c}{\multirow{2}{*}{\diagbox{Model}{Metrics}}}          & \multicolumn{4}{c}{Acceptance Prediction}       & \multicolumn{2}{c}{Score Prediction}        \\
\cmidrule(lr){2-5} \cmidrule(lr){6-7}
\multicolumn{1}{c}{\textbf{}} & \multicolumn{1}{c}{Accuracy} & \multicolumn{1}{c}{Precision} & \multicolumn{1}{c}{Recall} & \multicolumn{1}{c}{F1} & \multicolumn{1}{c}{MAE} & \multicolumn{1}{c}{MSE} \\
\midrule
LLaMA-3.1-8B (zero-shot)          & \bad{60.19\tiny $\pm$0.00}  & \bad{60.19\tiny $\pm$0.00}   & \bad{100.00\tiny $\pm$0.00} & \bad{75.13\tiny $\pm$0.00} & 1.961\tiny $\pm$0.055 & 5.488\tiny $\pm$0.287 \\
Qwen2.5-7B (zero-shot)           & \bad{60.19\tiny $\pm$0.00}  & \bad{60.19\tiny $\pm$0.00}   & \bad{100.00\tiny $\pm$0.00} & \bad{75.13\tiny $\pm$0.00} & 2.043\tiny $\pm$0.000 & 5.756\tiny $\pm$0.000 \\
LLaMA-3.1-8B (SFT)    & 63.23\tiny $\pm$0.33  & \textbf{73.66}\tiny $\pm$0.64   & 63.74\tiny $\pm$0.72  & 68.35\tiny $\pm$0.12 & \underline{1.141}\tiny $\pm$0.030 & \underline{2.290}\tiny $\pm$0.108 \\
Qwen2.5-7B (SFT)     & 62.01\tiny $\pm$0.00   & 72.25\tiny $\pm$0.01    & 62.18\tiny $\pm$0.00   & 66.83\tiny $\pm$0.00  & 1.182\tiny $\pm$0.000   & 2.374\tiny $\pm$0.000   \\
\midrule
SEA-E & \underline{66.24}\tiny $\pm$1.42  & 66.69\tiny $\pm$0.68   & \textbf{91.71}\tiny $\pm$0.82  & \textbf{77.23}\tiny $\pm$0.74 & 1.157\tiny $\pm$0.044 & 2.304\tiny $\pm$0.176 \\
LLaMA-OpenReviewer-8B         & 59.76\tiny $\pm$0.21  & 70.92\tiny $\pm$0.25   & 53.59\tiny $\pm$1.05  & 61.05\tiny $\pm$0.77 & 1.197\tiny $\pm$0.012 & 2.413\tiny $\pm$0.021 \\
DeepReviewer-14B  & \textbf{67.89}\tiny $\pm$1.99  & 65.75\tiny $\pm$1.31   & \underline{84.43}\tiny $\pm$1.87  & \underline{73.93}\tiny $\pm$1.51 & \textbf{1.104}\tiny $\pm$0.021 & \textbf{2.018}\tiny $\pm$0.087 \\
CycleReviewer-8B  & 54.18\tiny $\pm$0.27  & \underline{73.44}\tiny $\pm$1.96   & 33.25\tiny $\pm$2.01  & 45.73\tiny $\pm$1.50 & 1.321\tiny $\pm$0.008 & 2.941\tiny $\pm$0.015 \\
\bottomrule
\end{tabular}
}

\vspace{-1mm}

\end{table}

\begin{table}[ht]
\centering
\caption{ \label{tab:review_exp_gene}
    Results comparison between open-source LLMs and baselines on review generation.
}

\vspace{-2mm}

% \scriptsize
\footnotesize
{

\begin{tabular}{lrrrrrr}
\toprule
\multicolumn{1}{c}{\multirow{2}{*}{\diagbox{Model}{Metrics}}}  & \multicolumn{1}{c}{\multirow{2}{*}{BLEU}} & \multicolumn{1}{c}{\multirow{2}{*}{ROUGE-L}} & \multicolumn{2}{c}{BERTScore}        & \multicolumn{1}{c}{\multirow{2}{*}{EmbedCos}} \\% & \multicolumn{1}{c}{\multirow{2}{*}{LLM Judge}} \\
\cmidrule(lr){4-5}
\multicolumn{1}{c}{} & \multicolumn{1}{c}{}     & \multicolumn{1}{c}{}         & \multicolumn{1}{c}{Precision} & \multicolumn{1}{c}{Recall} & \multicolumn{1}{c}{}        \\%  & \multicolumn{1}{c}{}           \\
\midrule
LLaMA-3.1-8B (zero-shot) & 1.52\tiny $\pm$0.02 & 16.29\tiny $\pm$0.02  & 59.76\tiny $\pm$0.03& \underline{58.60}\tiny $\pm$0.01& 0.460\tiny $\pm$0.002 \\% & 7.512\tiny $\pm$0.008   \\
Qwen2.5-7B (zero-shot)  & 1.37\tiny $\pm$0.00 & 16.17\tiny $\pm$0.00  & \textbf{60.08}\tiny $\pm$0.00& \textbf{60.21}\tiny $\pm$0.00& 0.451\tiny $\pm$0.000 \\% & 7.683\tiny $\pm$0.009   \\
LLaMA-3.1-8B (SFT)       & \textbf{2.50}\tiny $\pm$0.18 & \textbf{17.92}\tiny $\pm$0.52  & \underline{59.88}\tiny $\pm$0.33 & 53.05\tiny $\pm$0.42 & \textbf{0.730}\tiny $\pm$0.094 \\% & 7.776\tiny $\pm$0.000   \\
Qwen2.5-7B (SFT)        & \underline{1.96}\tiny $\pm$0.01 & \underline{17.25}\tiny $\pm$0.02  & 58.34\tiny $\pm$0.02 & 52.13\tiny $\pm$0.01 & \underline{0.587}\tiny $\pm$0.002 \\% & 7.231\tiny $\pm$0.000   \\
\midrule
SEA-E & 1.25\tiny $\pm$0.19 & 15.72\tiny $\pm$0.61  & 59.83\tiny $\pm$0.67& 57.87\tiny $\pm$0.56& 0.385\tiny $\pm$0.101 \\% & 7.442\tiny $\pm$0.019   \\
LLaMA-OpenReviewer-8B& 1.49\tiny $\pm$0.05 & 15.90\tiny $\pm$0.31  & 59.38\tiny $\pm$0.48& 54.39\tiny $\pm$0.60& 0.444\tiny $\pm$0.008 \\% & 6.998\tiny $\pm$0.016   \\
DeepReviewer-14B     & 0.62\tiny $\pm$0.12 & 8.74\tiny $\pm$0.30   & 53.64\tiny $\pm$0.54& 57.77\tiny $\pm$0.78& 0.354\tiny $\pm$0.077 \\% & 7.592\tiny $\pm$0.071   \\
CycleReviewer-8B     & 0.68\tiny $\pm$0.01 & 11.93\tiny $\pm$0.02  & 54.97\tiny $\pm$0.12& 53.84\tiny $\pm$0.14& 0.487\tiny $\pm$0.024 \\% & 7.015\tiny $\pm$0.014   \\  
\bottomrule
\end{tabular}
}
\end{table}

\vpara{Rebuttal-Discussion Conversation. }
Tab.~\ref{tab:rebuttal_exp_1} and Tab.~\ref{tab:rebuttal_exp_2} show the experimental results for the rebuttal-discussion conversation task using two categories of evaluation metrics: semantic similarity and LLM-as-judge scores.
For the lexical and semantic similarity metrics (Tab.~\ref{tab:rebuttal_exp_1}), the fine-tuned LLaMA-3.1-8B model achieves the best performance on four of the five metrics. Moreover, both the fine-tuned LLaMA-3.1-8B and Qwen2.5-7B consistently outperform their zero-shot versions on nearly all metrics, indicating that training on our dataset effectively enhances the LLM’s multi-turn conversation ability in the rebuttal scenario.
As shown in Tab.~\ref{tab:rebuttal_exp_2}, similar trends are observed in the LLM-as-judge evaluation. 
Across the five evaluation aspects judged by the LLM, the fine-tuned LLaMA-3.1-8B also achieves the best overall performance, and the fine-tuned two models generally surpass their zero-shot versions.
These results clearly demonstrate the effectiveness of our training data in improving the capabilities of language models in author-reviewer discussion scenarios.

\begin{table}[ht]
\centering
\caption{ \label{tab:rebuttal_exp_1}
    Results on the lexical and semantic similarity metrics for rebuttal-discussion conversation.
}

\vspace{-2mm}

% \scriptsize
\footnotesize
\setlength{\tabcolsep}{1.7mm} %列距离
{

\begin{tabular}{lrrrrr}
\toprule
\multicolumn{1}{c}{\multirow{2}{*}{\diagbox{Model}{Metrics}}}  & \multicolumn{1}{c}{\multirow{2}{*}{BLEU}} & \multicolumn{1}{c}{\multirow{2}{*}{ROUGE-L}} & \multicolumn{2}{c}{BERTScore}        & \multicolumn{1}{c}{\multirow{2}{*}{EmbedCos}}\\ %& \multicolumn{1}{c}{\multirow{2}{*}{quality}} & \multicolumn{1}{c}{\multirow{2}{*}{constructive}} & \multicolumn{1}{c}{\multirow{2}{*}{accuracy}}  & \multicolumn{1}{c}{\multirow{2}{*}{completeness}} & \multicolumn{1}{c}{\multirow{2}{*}{clarity}}\\
\cmidrule(lr){4-5}
\multicolumn{1}{c}{} & \multicolumn{1}{c}{}     & \multicolumn{1}{c}{}         & \multicolumn{1}{c}{Precision} & \multicolumn{1}{c}{Recall} & \multicolumn{1}{c}{}        \\
\midrule
LLaMA-3.1-8B (zero-shot) & 1.23\tiny $\pm$0.01 & \underline{13.13}\tiny $\pm$0.02  & \underline{52.86}\tiny $\pm$0.03& \textbf{59.72}\tiny $\pm$0.02& 0.516\tiny $\pm$0.001 \\
Qwen2.5-7B (zero-shot)  & 1.05\tiny $\pm$0.00 & 11.74\tiny $\pm$0.00  & 48.96\tiny $\pm$0.00& \underline{58.42}\tiny $\pm$0.00& 0.454\tiny $\pm$0.000 \\
LLaMA-3.1-8B (SFT)       & \textbf{2.07}\tiny $\pm$0.32 & \textbf{14.78}\tiny $\pm$0.61  & \textbf{54.15}\tiny $\pm$0.47 & 58.34\tiny $\pm$0.54 & \textbf{0.889}\tiny $\pm$0.097  \\
Qwen2.5-7B (SFT)        & \underline{1.46}\tiny $\pm$0.01 & 12.36\tiny $\pm$0.02  & 49.38\tiny $\pm$0.02 & 58.09\tiny $\pm$0.01 & \underline{0.612}\tiny $\pm$0.003  \\
\bottomrule
\end{tabular}
}

\end{table}

\begin{table}[ht]
\centering
\caption{ \label{tab:rebuttal_exp_2}
    Results on the LLM-as-judge metrics for rebuttal-discussion conversation.
}

\vspace{-2mm}

% \scriptsize
\footnotesize
\setlength{\tabcolsep}{1.7mm} %列距离
{

\begin{tabular}{lcccccc}
\toprule
\multicolumn{1}{c}{\multirow{2}{*}{\diagbox{Model}{Metrics}}} & \multicolumn{1}{c}{\multirow{2}{*}{Quality}} & \multicolumn{1}{c}{\multirow{2}{*}{Constructiveness}} & \multicolumn{1}{c}{\multirow{2}{*}{Accuracy}}  & \multicolumn{1}{c}{\multirow{2}{*}{Completeness}} & \multicolumn{1}{c}{\multirow{2}{*}{Clarity}}\\

\multicolumn{1}{c}{} & \multicolumn{1}{c}{}     & \multicolumn{1}{c}{}         & \multicolumn{1}{c}{} & \multicolumn{1}{c}{} & \multicolumn{1}{c}{} & \multicolumn{1}{c}{}        \\
\midrule
LLaMA-3.1-8B (zero-shot) & 6.936\tiny $\pm$0.013 & 8.380\tiny $\pm$0.013  & \underline{8.032}\tiny $\pm$0.003& 7.133\tiny $\pm$0.005& \underline{8.297}\tiny $\pm$0.011 \\
Qwen2.5-7B (zero-shot)  & 6.861\tiny $\pm$0.000 & 8.373\tiny $\pm$0.000  & 7.677\tiny $\pm$0.000& 6.990\tiny $\pm$0.000& 7.944\tiny $\pm$0.000 \\
LLaMA-3.1-8B (SFT)       & \textbf{7.347}\tiny $\pm$0.022 & \textbf{8.603}\tiny $\pm$0.014  & \textbf{8.229}\tiny $\pm$0.003 & \textbf{7.210}\tiny $\pm$0.006 & \textbf{8.322}\tiny $\pm$0.009 \\
Qwen2.5-7B (SFT)        & \underline{7.228}\tiny $\pm$0.001 & \underline{8.521}\tiny $\pm$0.002  & 7.798\tiny $\pm$0.002 & \underline{7.183}\tiny $\pm$0.001 & 7.936\tiny $\pm$0.001  \\
\bottomrule
\end{tabular}
}

\vspace{-1mm}

\end{table}

\section{Related Work}

\vpara{Static Review Datasets.}
Initially, datasets in the field of automated paper review primarily collected static data such as manuscript drafts and review comments for review process analysis, acceptance rate prediction, and review generation tasks.
\citet{kang2018dataset} present the first public dataset of scientific peer reviews available for research purposes, and also propose two NLP tasks, acceptance prediction and aspect scores prediction based on their dataset.
\citet{dycke2022nlpeer} introduce an ethically sourced multi-domain corpus of more than 5k papers and 11k review reports from five different venues.
% They also establish a unified data representation and augment previous peer review datasets to include parsed and structured paper representations, rich metadata and versioning information.
\citet{zhang2022investigating} conduct a thorough and rigorous study on fairness disparities in peer review with the help of LLMs, observing that the level of disparity varies and textual features are essential in reducing biases in predictive modeling. 
% Their database offers opportunities for exploring new natural language processing (NLP) methods to enhance the understanding of peer review mechanisms.
\citet{gao2024reviewer2} propose an efficient two-stage review generation framework and generate a large-scale review dataset that annotated with aspect prompts.
\citet{zhou2024llm} first evaluate GPT-3.5 and GPT-4 on the score prediction task and the review generation task. They also propose a dataset comprising 197 review-revision multiple-choice questions (RR-MCQ) with detailed labels from the review-rebuttal forum in ICLR 2023. 
\citet{zhu2025deepreview} introduce DeepReview, a multi-stage framework designed to emulate expert reviewers by incorporating structured analysis, literature retrieval, and evidence-based argumentation.
However, although the above works have made progress in scholarly peer review,
most of these datasets could not guarantee the consistency between their paper content and those under review, nor did they include rebuttals, making it difficult to support the research about the reviewers-authors discussions.

\vpara{Rebuttal-included Datasets.}
Different from the aforementioned works, some other researchers begin to incorporate information from the rebuttal stage into peer review datasets.
\citet{kennard2021disapere} synthesize label sets from prior work and extend them to include fine-grained annotation of the rebuttal sentences, characterizing their context in the review and the authors' stance towards review arguments.
\citet{jin2024agentreview} introduce AgentReview, the first LLM-based peer review simulation framework, which effectively disentangles the impacts of multiple latent factors and addresses the privacy issue.
\citet{wu2022incorporating} present a novel generation model that is capable of explicitly modeling the complicated argumentation structure from not only arguments between the reviewers and the authors, but also the inter-reviewer discussions.
% \citet{tan2024peer} construct a comprehensive dataset to facilitate the applications of LLMs for multi-turn dialogues, effectively simulating the complete peer-review process. 
However, although these datasets include rebuttal content, they still suffer from several limitations, such as the issue of original manuscript versions, limited data diversity and scale, or the unreliability of simulated data.

To address the limitations mentioned above, our \useVal{model} dataset introduces several key advancements. 
First, we make sure that all the 19,926 papers in our dataset are their initial submission versions, ensuring version reliability and data consistency.
Second, our dataset is the largest real-world peer review dataset, which features the widest range of conferences and the most complete review stages.
Finally, beyond the trivial static review paradigm, we propose to treat the rebuttal and discussion data as a multi-turn conversation task between reviewers and authors, paving the way for the training and evaluation of dynamic, interactive LLM-based reviewing assistants.

\section{Conclusion}

% In this work, we present the largest real-world peer review dataset named \useVal{model}, to support the training and evaluation of both review and rebuttal abilities of language models.
% With the broadest coverage of conferences and the most comprehensive inclusion of review stages,
% it is the first consistency-ensured dataset to support rebuttal tasks in a multi-turn conversation paradigm. 
% Our \useVal{model} dataset not only allows traditional static tasks such as score prediction, review generation, and computational analysis, but also enables the training of interactive, chat-based models for further rebuttal and discussion. 
% Our goal is to assist authors in self-evaluating and refining their work prior to submission, ultimately helping to alleviate the review burden within the AI research community.

In this work, we present \useVal{model}, the largest real-world peer review dataset that ensures data consistency, to support the model training and evaluation on tasks related to both review and rebuttal.
% With the broadest conference coverage and the most comprehensive representation of review stages, 
% it is the first consistency-ensured dataset to support rebuttal tasks within a multi-turn conversational paradigm.
Our \useVal{model} dataset facilitates not only traditional static tasks,
% such as score prediction, review generation, and computational analysis, 
but also the training of interactive, dialogue-based models for further rebuttal and discussion.
In this way, not only can the review burden on reviewers be reduced, but authors are also better able to self-evaluate and improve their manuscripts before submission, ultimately helping to ease the overall reviewing pressure in the AI research community.
This not only reduces the review burden on reviewers, but also helps authors self-evaluate and improve their manuscripts before submission, easing the overall review pressure in the AI community.

\vpara{Limitations. }
% By collecting and organizing as much public peer review data as possible, 
Although our \useVal{model} dataset represents the most comprehensive resource in the peer review field,
since existing review works mainly focus on textual content, our experiments are limited to the textual and tabular components of papers, excluding visual elements such as figures (even though they are included in the released dataset).
As the field advances, our future work will benefit from the integration of vision-language models,
% This would allow for richer semantics that combine textual and visual modalities, and ultimately support a wider range of applications in automated scholarly review.
which will offer richer semantics combining textual and visual modalities and ultimately support wider applications in automated academic review.

\newpage
\bibliography{neurips_2025.bib}
\bibliographystyle{unsrtnat}

\newpage
\appendix

\section{Details of the Violin Plot}\label{app:violin_plot}

\begin{figure}[ht]
    \centering
    \includegraphics[width=\linewidth / 4 * 3]{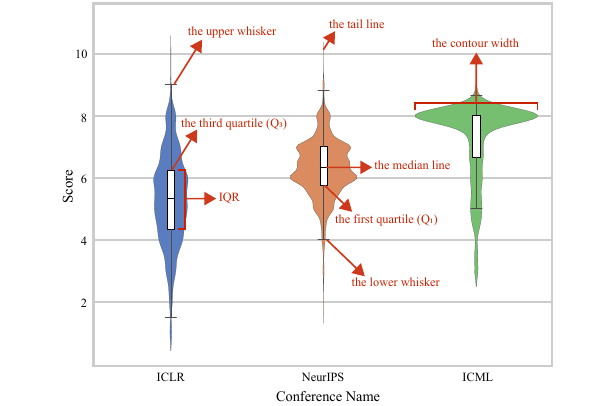}
    \caption{
    \textbf{Details of the Violin Plot.}
    The upper and lower whiskers represent the maximum and minimum observed values.
    The third quartile ($\text{Q}_3$) is the value below which 75\% of the data falls.
    The first quartile ($\text{Q}_1$) is the value below which 25\% of the data falls.
    IQR represents the distribution of the middle 50\% of the data.
    The tail line can extend beyond the data boundaries, reflecting the smoothness of the curve.
    The median line's position varies with the data distribution.
    The contour width represents the density of the number of papers.
    }
    \label{fig:appendix_Violin_Plot}
\end{figure}

To illustrate the (normalized) review score distributions of each conference, we utilize the violin plot to conduct a statistical analysis of the normalized scores in Fig.~\ref{fig:statistic_dataset}(f). 
Here we introduce more details of the violin plot.
The violin plot is a statistical graphic that combines the box plot and the kernel density plot, allowing for a better display of the data distribution. As shown in Fig.~\ref{fig:appendix_Violin_Plot}, each violin represents the score distribution of a conference.

\vpara{Kernel density estimation.}
The wider the contour, the more papers are assigned this score. The smoothness of the curve is controlled by the bandwidth. The larger the value, the smoother the curve. This also results in tail lines that go beyond the data boundaries, with no real data points typically present in them. The kernel density estimation function is as follows:
\begin{equation}
\hat{f}(x) = \frac{1}{n h} \sum_{i=1}^{n} K\left( \frac{x - x_i}{h} \right)
\end{equation}
where $\hat{f}(x)$ denotes the estimated density function value at position $x$;
$n$ denotes the sample size;
$h$ denotes the bandwidth, which controls the smoothness of the kernel function;
$x_i$ denotes the $i$-th sample point;
$K(\cdot)$ denotes the kernel function, usually a symmetric probability density function. 
We use Seaborn\footnote{\url{https://seaborn.pydata.org/}} to plot the violin plot, which defaults to the Gaussian kernel. $x$ represents the estimation point and $x_i$ represents the true sample point. The kernel function is as follows:
\begin{equation}
    K\left( \frac{x - x_i}{h} \right) = \frac{1}{\sqrt{2\pi}} \exp\left( - \frac{1}{2} \left( \frac{x - x_i}{h} \right)^2 \right)
\end{equation}

\vpara{Boxplot.}  
The box plot consists of the box, the median line, and the upper and lower whiskers. The upper boundary of the box is the third quartile ($\text{Q}_3$), which indicates that 75\% of the data points are below this value; the lower boundary of the box is the first quartile ($\text{Q}_1$), which indicates that 25\% of the data points are below this value. The total height of the box is the Interquartile Range ($\text{IQR} = \text{Q}_3 - \text{Q}_1$), which represents the spread of the middle 50\% of the data.

The median line ($\text{Q}_2$) is usually located in the center of the box, but if the data distribution is not balanced, the position of the line may shift. When most of the data is concentrated in the lower values, the median line will be closer to the lower boundary; when most of the data is concentrated in the higher values, the median line will be closer to the upper boundary.
In Fig.~\ref{fig:appendix_Violin_Plot}, for ICML, the median line is near the third quartile because most of the papers have scores around 8.

The upper and lower whiskers represent the maximum and minimum observed values, and they are typically used to identify outliers. The formulas are as follows:
\begin{equation}
\label{eq:lower_whisker}
\text{Lower Whisker} = \max\left(\min(X), \text{Q}_1 - 1.5 \times \text{IQR}\right)
\end{equation}
\begin{equation}
\label{eq:upper_whisker}
\text{Upper Whisker} = \min\left(\max(X), \text{Q}_3 + 1.5 \times \text{IQR}\right)
\end{equation}

When analyzing the distribution of review scores, most of the data falls between the upper and lower whiskers. As shown in equations~\ref{eq:lower_whisker} and~\ref{eq:upper_whisker}, neither the upper whisker nor the lower whisker represents the actual maximum or minimum values. Since very few papers have such low scores, these scores can be considered outliers and placed below the lower whisker.
From Fig.~\ref{fig:appendix_Violin_Plot}, we can see that the curve for ICLR is smoother compared to the other two conferences, with fewer peaks. Therefore, the paper score distribution for ICLR is the most evenly spread.

\section{Details of the Peer Review Methods in the Experiment}\label{app:baselines}

In addition to the open-source LLMs like LLaMA-3.1-8B-Inst and Qwen2.5-7B-Inst, we further evaluate some models specifically designed for paper review scenarios on our test dataset, including SEA-E~\citep{yu2024automated}, LLaMA-OpenReviewer-8B~\citep{idahl2024openreviewer}, DeepReviewer-14B~\citep{zhu2025deepreview}, and CycleReviewer-8B~\citep{weng2024cycleresearcher}. 
The details of these methods are given below:
\begin{itemize}
    \item SEA-E~\citep{yu2024automated}:
    It is the evaluation model from an automated paper reviewing framework named SEA, which comprises of three modules: Standardization, Evaluation, and Analysis. SEA-E utilizes the standardized data that is integrated from multiple reviews for fine-tuning, enabling it to generate constructive reviews.
    
    \item LLaMA-OpenReviewer-8B~\citep{idahl2024openreviewer}:
    The OpenReviewer is an open-source system for generating high-quality peer reviews of machine learning and AI conference papers. Its core is LLaMA-OpenReviewer-8B, an 8B parameter language model specifically fine-tuned on 79,000 expert reviews from top conferences, which produces considerably more critical and realistic reviews compared to general-purpose LLMs like GPT-4 and Claude-3.5.
    
    \item DeepReviewer-14B~\citep{zhu2025deepreview}:
    The DeepReview is a multi-stage framework designed to emulate expert reviewers by incorporating structured analysis, literature retrieval, and evidence-based argumentation. Using DeepReview-13K, a curated dataset with structured annotations, DeepReviewer-14B is trained and outperforms CycleReviewer-70B with fewer tokens. In its best mode, DeepReviewer-14B achieves win rates of 88.21\% and 80.20\% against GPT-o1 and DeepSeek-R1 in evaluations.
    
    \item CycleReviewer-8B~\citep{weng2024cycleresearcher}:
    The author explores the feasibility of using open-source post-trained LLMs as autonomous agents capable of performing the full cycle of automated research and review, from literature review and manuscript preparation to peer review and paper refinement. In this iterative preference training framework, CycleReviewer simulates the peer review process, providing iterative feedback via reinforcement learning. 
    
\end{itemize}

\section{Details of the Evaluation Metrics}\label{app:eval_metrics}
\vpara{BLEU.}
BLEU (Bilingual Evaluation Understudy) mainly measures the similarity between texts using n-gram overlap. An n-gram is a sequence of $n$ consecutive words or characters in a text. The n-gram overlap refers to how many n-grams in the generated text match with those in the reference text, reflecting local similarity. For n-grams, a smaller $n$ makes it easier to match, but it loses context; a larger $n$ makes it harder to match, but better reflects sentence structure and includes more contextual information. Therefore, we need to combine the n-gram precision for different values of $n$, usually by calculating the geometric mean, and add a penalty factor (BP) to prevent the generated text from being too short. The formulas are as follows:
\begin{equation}
P_n = \frac{
  \sum_{C \in \text{Candidates}} \sum_{\text{n-gram} \in C} \text{Count}_{\text{clip}}(\text{n-gram})
}{
  \sum_{C \in \text{Candidates}} \sum_{\text{n-gram} \in C} \text{Count}(\text{n-gram})
}
\end{equation}

\begin{equation}
\text{BP} = 
\begin{cases}
1, & \text{if } c > r \\
\exp\left(1 - \frac{r}{c} \right), & \text{if } c \leq r
\end{cases}
\end{equation}

\begin{equation}
\text{BLEU} = \text{BP} \cdot \exp \left( \Sigma_{n=1}^{N} w_n \log P_n \right)
\end{equation}

Here BP stands for brevity penalty, which is used to penalize overly short outputs; $c$ denotes the length of the generated text; $r$ denotes the length of the reference text; $w_n$ denotes the weight of the n-gram (usually $\frac{1}{n}$). 
It’s important to note that BLEU is not sensitive to synonyms or grammatical variations. It doesn’t work well for evaluating single sentences and is more suitable for evaluating longer texts or entire documents, which matches the tasks about review generation we focus on.

\vpara{ROUGE.}
ROUGE (Recall-Oriented Understudy for Gisting Evaluation) is commonly used to evaluate how much of the reference text is covered by the output from tasks like text summarization and question answering. It has several variants, such as ROUGE-N (based on n-gram recall), ROUGE-L (based on the Longest Common Subsequence, or LCS), ROUGE-W (weighted LCS), and ROUGE-S / ROUGE-SU (based on skip-bigram matches).

Taking ROUGE-L as an example, the LCS is the longest sequence of words that appear in both texts in the same order, but not necessarily next to each other.
In our experiment, we report the ROUGE-L F1 score, which is the harmonic mean of ROUGE-L Precision and ROUGE-L Recall.
\begin{equation}
\text{ROUGE-L}_\text{Precision} = \frac{L}{m}, \quad \text{ROUGE-L}_\text{Recall} = \frac{L}{n}
\end{equation}

\begin{equation}
\text{ROUGE-L}_{\text{F}_\beta} = 
\frac{
(1 + \beta^2) \cdot \text{ROUGE-L}_\text{Precision} \cdot \text{ROUGE-L}_\text{Recall} }
{
\beta^2 \cdot \text{ROUGE-L}_\text{Precision} + \text{ROUGE-L}_\text{Recall}
}
\end{equation}
Here $L$ is the length of the LCS, $m$ is the length of the reference text, and $n$ is the length of the generated text.
When there are multiple reference texts, ROUGE compares the generated text with each one, calculates a ROUGE-L score for each pair, and then selects the highest ROUGE-L score as the final result.

ROUGE has similar limitations to BLEU. Since both are based on lexical matching at the surface level, they cannot effectively evaluate semantic similarity.
Therefore, we often further use semantic evaluation metrics, such as BERTScore and the cosine similarity of embeddings (EmbedCos).

\vpara{BERTScore.}
BERTScore uses BERT to extract word embeddings and measures how well two texts match by comparing the similarity between their word vectors. First, BERT embeddings are generated separately for $r_i$ (the reference) and $c_i$ (the generated text). Then, using the word embeddings $r_i$ and $c_i$, pairwise cosine similarities are calculated to form a similarity matrix $S_{ij}$ of size $n \times m$:
\begin{equation}
S_{ij} = \cos(\vec{c}_i, \vec{r}_j) = \frac{\vec{c}_i \cdot \vec{r}_j}{\|\vec{c}_i\| \cdot \|\vec{r}_j\|}
\end{equation}
After obtaining $S_{ij}$, we can calculate the Precision and Recall of BERTScore.
Precision is computed by finding, for each word in the candidate sentence, the most similar word in the reference sentence, and then taking the average of these maximum similarities.
Recall is calculated by doing the same in reverse: for each word in the reference sentence, find the most similar word in the candidate sentence, and then take the average.
\begin{equation}
\text{Precision} = \frac{1}{n} \sum_{i=1}^{n} \max_{1 \leq j \leq m} S_{ij}, \quad
\text{Recall} = \frac{1}{m} \sum_{j=1}^{m} \max_{1 \leq i \leq n} S_{ij}
\end{equation}
BERTScore does not require exact word matching, allowing it to recognize synonyms to some extent and providing some level of contextual awareness. However, it is entirely based on word-level matching and does not take sentence structure or grammatical order into account. It requires a large amount of computation, and its results are harder to interpret compared to BLEU and ROUGE. 
In our work, we use a 12-layer DeBERTa-large-MNLI for the strong semantic understanding capability.

\vpara{EmbedCos.}
EmbedCos refers to the cosine similarity of embeddings. We use the \texttt{sentence-transformers/all-mpnet-base-v2} model to calculate EmbedCos, because compared to other base Transformer models like BERT and RoBERTa, MPNet is more efficient and can generate high-quality sentence embeddings in less time. 
This model focuses on sentence-level embeddings and optimizes the semantic distance between sentences, so similar sentences are closer together in the vector space, and different sentences are farther apart.

The calculation process of EmbedCos can be divided into two steps:
First, a set of sentences is tokenized and passed into MPNet. The embeddings obtained here represent the entire sentence's semantic meaning, unlike BERTScore, which uses individual words. 
% The model internally uses mean pooling, and its formula is $\vec{s} = \frac{1}{n} \sum_{i=1}^{n} \vec{t}_i$.
Second, based on the obtained sentence embeddings, the EmbedCos is computed as the cosine similarity between sentences:
\begin{equation}
\text{EmbedCos}(\vec{u}, \vec{v}) = \frac{\vec{u} \cdot \vec{v}}{\|\vec{u}\| \cdot \|\vec{v}\|} \in [-1, 1]
\end{equation}
Here $\vec{u} \cdot \vec{v}$ is the dot product of the vectors, and $||\vec{u}||$ is the L2 norm of $\vec{u}$.

As shown in Tab.~\ref{tab/Evaluation_metrics_comparison}, we show the comparison between the four metrics: BLEU, ROUGE-L, BERTScore, and EmbedCos.
\begin{table}[H]
\centering 
\caption{Comparison of the evaluation metrics.}
\label{tab/Evaluation_metrics_comparison}
\footnotesize
\setlength{\tabcolsep}{1.7mm} 
\renewcommand{\arraystretch}{1.5} 
\begin{tabular}{>{\centering\arraybackslash}m{1.85cm} 
                >{\centering\arraybackslash}m{1.9cm} 
                >{\centering\arraybackslash}m{2cm} 
                >{\centering\arraybackslash}m{3cm} 
                >{\centering\arraybackslash}m{3cm}}
\toprule
\textbf{Metric} & \textbf{BLEU} & \textbf{ROUGE-L} & \textbf{BERTScore} & \textbf{EmbedCos} \\ \hline
\textbf{Evaluation Method} & N-gram Precision & LCS & Token-level Semantic Similarity (BERT) & Cosine Similarity of Sentence Embeddings \\ \hline
\textbf{Semantic Focus} &  &  & \checkmark & \checkmark \\ 
\midrule
\textbf{Model Dependency} &  &  & \checkmark & \checkmark \\ \midrule
\textbf{Evaluation Granularity} & N-gram & LCS & Token-level Similarity Aggregation & Sentence-level Embedding Similarity \\
\bottomrule
\end{tabular}
\end{table}

\section{Details of the LLM as Judge}\label{app:rebuttal_llm_judge}
In the rebuttal-discussion conversation task, we use LLaMA-3.1-8B-Inst to evaluate the quality of generated dialogue statements. 
Here we instruct the judge to focus on five aspects to evaluate the results generated by the reviewer model, including accuracy, clarity, constructiveness, completeness, and quality.
\begin{itemize}
    \item Accuracy is selected to ensure the correctness of the model's output, making sure the generated responses align logically with the input questions or context and avoid factual errors.
    \item Clarity assesses whether the generated statements are expressed fluently and understandably, avoiding ambiguity or redundancy, thereby reflecting the model's language organization capability.
    \item Constructiveness focuses on the relevance and practical value of the generated content, such as whether it can provide effective improvement suggestions for contentious points, demonstrating the usefulness of the dialogue.
    \item Completeness measures whether the response covers key issues or arguments, avoiding the omission of important information, especially in academic rebuttals where comprehensive responses to critiques are essential.  
    \item Quality serves as an overall evaluation criterion, integrating all the above dimensions to grade the generated content.
    
\end{itemize}  

Among these metrics, accuracy and clarity emphasize the model's output quality at a technical level; constructiveness emphasizes dialogue relevance at a practical level; completeness focuses on comment coverage at a content level; and quality acts as a global indicator.

The instruction we used to guide the LLaMA-3.1-8B-Inst as a judge is shown as below:

\begin{tcolorbox}[colback=white, left=5mm]
\begin{verbatim}
You are an expert evaluator. 
Given the gold reference answer and a candidate answer, score the 
candidate’s quality, constructive, accuracy, completeness and 
clarity on a scale of 1‑10.
- Quality: Scores overall depth, logic, and usefulness—low for 
shallow/chaotic content, high for insightful and valuable input.
- Constructive: Measures whether the comment offers solutions or 
fosters discussion—low for pure criticism, high for actionable 
feedback.
- Accuracy: Rates factual/logical correctness—low for 
errors/misleading claims, high for well-supported and precise 
statements.
- Completeness: Assesses coverage of key points—low for major 
omissions, high for thorough and detailed analysis.
- Clarity: Judges coherence and readability—low for 
confusing/verbose language, high for concise and well-structured 
delivery.
Please apply stricter grading criteria to reduce the proportion 
of high scores and ensure a reasonable score distribution. Only 
exceptionally outstanding performances should receive high scores
, average performances should receive moderate scores, and poor 
performances should receive low scores. Be more objective and 
conservative in your grading.
Respond strictly as JSON, do not provide any other content:
{
    "quality": <int>,
    "constructive": <int>,
    "accuracy": <int>,
    "completeness": <int>,
    "clarity": <int>
}
\end{verbatim}
\end{tcolorbox}

% \newpage
% \input{checklist}

\end{document}